\documentclass[utf8]{frontiersSCNS} 

\usepackage{url,hyperref,lineno,microtype,subcaption}
\usepackage[onehalfspacing]{setspace}

\usepackage{multirow}

\def\keyFont{\fontsize{8}{11}\helveticabold }
\def\firstAuthorLast{Allred {et~al.}} 
\def\Authors{Jason M. Allred\,$^{1,*}$, Steven J. Spencer\,$^{1}$, Gopalakrishnan Srinivasan\,$^{1}$, and Kaushik Roy\,$^{1}$}
\def\Address{$^{1}$Nanoelectronics Research Laboratory, Electrical and Computer Engineering Department, Purdue University, West Lafayette, IN, USA
}
\def\corrAuthor{Jason M. Allred}

\def\corrEmail{allredj@purdue.edu}

\begin{document}
\onecolumn
\firstpage{1}

\title[Explicitly Trained Spiking Sparsity]{Explicitly Trained Spiking Sparsity in Spiking Neural Networks with Backpropagation} 

\author[\firstAuthorLast ]{\Authors} 
\address{} 
\correspondance{} 

\extraAuth{}

\maketitle

\begin{abstract}

\section{}
Spiking Neural Networks (SNNs) are being explored for their potential energy efficiency resulting from sparse, event-driven computations. Many recent works have demonstrated effective backpropagation for deep Spiking Neural Networks (SNNs) by approximating gradients over discontinuous neuron \textit{spikes} or firing events. A beneficial side-effect of these surrogate gradient spiking backpropagation algorithms is that the spikes, which trigger additional computations, may now themselves be directly considered in the gradient calculations. We propose an explicit inclusion of spike counts in the loss function, along with a traditional error loss, causing the backpropagation learning algorithms to optimize weight parameters for both accuracy \textit{and} spiking sparsity. As supported by existing theory of over-parameterized neural networks, there are many solution states with effectively equivalent accuracy. As such, appropriate weighting of the two loss goals during training in this multi-objective optimization process can yield an improvement in spiking sparsity without a significant loss of accuracy. We additionally explore a simulated annealing-inspired loss weighting technique to increase the weighting for sparsity as training time increases. Our preliminary results on the Cifar-10 dataset show up to 70.1\% reduction in spiking activity with iso-accuracy compared to an equivalent SNN trained only for accuracy and up to 73.3\% reduction in spiking activity if allowed a trade-off of 1\% reduction in classification accuracy.

\tiny
 \keyFont{ \section{Keywords:} spiking neural networks, sparsity, spiking backpropagation, energy efficiency, surrogate gradients} 
\end{abstract}

\section{Introduction}

Spiking Neural Networks (SNNs) are a type of Artificial Neural Network (ANN) being explored in machine learning in part for their potential energy efficiency benefits due to the inherent computational sparsity that comes from event-driven computation (\cite{han2017deepspikingenergy}). The computational energy consumed in a spiking network during inference is highly correlated with the number of spikes that occur because each spike at a given neuron induces accumulation computations in each of that neuron's fan-out neurons as well as a membrane reset computation. Thus, reducing spiking activity is an important part of improving energy efficiency in an SNN.

Until recently, training large-scale SNNs for complicated datasets was a difficult task because discontinuous neuron activations are non-differentiable, preventing direct backpropagation. One of the first workarounds to this problem was converting a pre-trained, non-spiking ANN to an SNN (\cite{sengupta2019deepspikingconversion}). This approach allowed for competitive inference on an SNN for complicated tasks like ImageNet, but it failed to capture the energy efficiency benefits of sparsity. This failure was because the networks were trained in a highly precise, deterministic environment, and switching to the stochastic environment of an SNN reduces resolution at small time scales, requiring a larger inference time to accurately distinguish between close activation values. This larger inference time results in a significant number of spiking operations, limiting energy efficiency benefits.

However, more recent works have demonstrated effective methods at backpropagating directly in a spiking environment, e.g. \cite{Emre2019arXivSurrogateGradient}, \cite{lee2019spikingbackprop}. These methods approximate the gradients over the discontinuous spiking activations, allowing for backpropagation through a deep SNN. \cite{lee2019spikingbackprop} have shown that this method of spiking backpropagation significantly reduces the inference time required, and, with that, the total number of spikes and computations that occur per inference, further improving the energy efficiency.

A beneficial side effect of these approximate spiking gradient techniques is that spikes themselves may now be included in these surrogate backpropagation learning algorithms. Since spiking sparsity is an energy goal of SNNs, we propose including spiking activity directly in the loss function, explicitly training the SNNs to be more sparse in a multi-objective optimization process. We additionally explore a simulated annealing-inspired loss function, providing backpropagation with a dynamic weighting of the two optimization goals (accuracy and sparsity), which may help avoid local minima or solution states that significantly sacrifice accuracy. Our preliminary results show that compared to training only for accuracy, explicitly including spiking sparsity in the loss function can achieve around 60-70\% reduction in spiking activity on Cifar10 with on-par accuracy.

\section{Materials and Methods}

\subsection{Using Surrogate Gradients for Backpropagation in Spiking Neural Networks}

In SNNs, a spiking activation is often modeled as inducing an instantaneous weighted potentiation in the membrane potentials of fan-out neurons. \cite{Emre2019arXivSurrogateGradient} have analyzed the effectiveness of various ``surrogate'' or approximate gradients over spiking activations, including fast-sigmoid, linear, and exponential surrogates, and have developed open-source code for easy backpropagation in PyTorch using these surrogate gradients, called SpyTorch (\cite{spytorch}). These approximate gradients allow us to choose any criterion for the loss function, $L()$, based on final classification error, e.g. mean squared error, cross entropy, etc., and let the automatic software tools perform backpropagation.

\begin{equation}
    L_{classification} = criterion(output, target)
\end{equation}

\subsection{Adding Spiking Activity to the Loss Functions}

\cite{Poggio2018Theory} have shown that the global minima for over-parameterized networks often reside in flat valleys or basins within the optimization space. This means that many neighboring solutions have effectively equivalent accuracy. These flat regions provide flexibility. Our goal is to let the optimizer find solutions within those flat basins that have less spiking activity without compromising accuracy.

Being able to differentiate over spiking events enables gradient descent based on a loss that includes a measure of those spikes. The total loss, then, can be a combination of both the classification accuracy goal \textit{and} the spiking sparsity goal:

\begin{equation}
    L_{total} = L_{classification} + L_{sparsity}
    \label{eq:combined_loss}
\end{equation}

\begin{equation}
    L_{sparsity} = \sigma(spikeCount)
\end{equation}

where $\sigma()$ is a weighting function that scales the spike count loss to provide appropriate balancing between the two loss components in (\ref{eq:combined_loss}).

\subsection{The Sparsity Loss Function, $\sigma()$}

The most trivial approach for $\sigma()$ would be to use a constant scalar:

\begin{equation}
    \sigma_{constant}(spikeCount) = \sigma_0 * spikeCount
\end{equation}

We consider a potential problem from adding in a sparsity loss function. When we change the optimization topography, if the optimizer is constrained for sparsity too much, too early in the training process, the gradients in the new landscape may not allow the system to reach solution states that also reside in the basins of the classification landscape, causing a significant reduction or complete failure of the classification accuracy. So for the constant sparsity loss scaling function, $\sigma_0$ must be small enough to allow the classification loss to dominate the total gradient direction if classification accuracy is to be maintained. However, letting the classification loss dominate too much, even after reaching the basins, may fail to achieve the best sparsity.

We explore a potential solution to this problem, inspired by simulated annealing--allow the optimizer to disregard sparsity early in the training process and then slowly increase the constraints for spiking sparsity as training continues. This approach makes $\sigma()$ a function of training time, or more simply, the current training epoch. In addition to the constant sparsity loss function, we evaluate four annealing-inspired sparsity loss functions that increase the sparsity constraint over time. These include a linear increase (equation (\ref{eq:linear})) and a quadratic increase (equation (\ref{eq:quadratic})). The other two loss functions alternate between excluding and including the sparsity loss function throughout the training process, where the portion of epochs in which the sparsity loss is included increases linearly during training from zero-inclusion during the first epoch to always-included in the last epoch. The first of these alternating loss functions uses a constant $\sigma_0$ when it the sparsity loss is included (equation (\ref{eq:alternating})), and the other uses a linearly increasing $\sigma_0$ when it is included (equation (\ref{eq:alternating_linear})).

\begin{equation}
    \sigma_{linear}(spikeCount,n_{epoch}) = \sigma_0 * \frac{n_{epoch}}{N_{epochs}} * spikeCount
    \label{eq:linear}
\end{equation}

\begin{equation}
    \sigma_{quadratic}(spikeCount,n_{epoch}) = \sigma_0 * \frac{(n_{epoch})^2}{N_{epochs}} * spikeCount
    \label{eq:quadratic}
\end{equation}

\begin{equation}
    \sigma_{alternating}(spikeCount,n_{epoch}) = A(n_{epoch}) * \sigma_0 * spikeCount
    \label{eq:alternating}
\end{equation}

\begin{equation}
    \sigma_{alternating\_linear}(spikeCount,n_{epoch}) = A(n_{epoch}) * \sigma_0 *  \frac{n_{epoch}}{N_{epochs}} * spikeCount
    \label{eq:alternating_linear}
\end{equation}

where $A()$ is a binary value following the the alternating function discussed above.


\subsection{Experimental Methodology}

Experiments were conducted in PyTorch with SpyTorch using backpropagation with piece-wise linear surrogate gradients on the Cifar-10 dataset with the LeNet architecture. (This paper will be updated with results from larger and deeper architectures.) Cross entropy was chosen as the classification loss function.  For each of the sparsity loss functions discussed above, we performed a hyperparameter search to discover the largest $\sigma_0$ that still provides acceptable classification accuracy based on the validation set. We set aside 20\% of the training set as a validation set and trained with the remaining 80\%. Both the sparsity loss scaling constant, $\sigma_0$, and the number of training epochs were determined based on the validation results.

\section{Results}

Using the hyperparameters for each method that gave the best spiking sparsity with \textit{zero} increase in validation error, we report the testing results in Table \ref{bestresults_strict}. Using the hyperparameters for each method that gave the best spiking sparsity with less than 1\% reduction in validation classification accuracy, we report the testing results in Table \ref{bestresults}.

\begin{table}[h]
\caption{Accuracy and spiking activity results using hyperparameters that give the best spiking sparsity with no increase in validation error.}
\centering
\begin{tabular}{l|l|l|r|r|r|}
\cline{3-6}
\multicolumn{1}{c}{}                                &                                     & \multicolumn{2}{c|}{Best Val. Hyperparameters} & \multicolumn{2}{c|}{Testing Results}  \\ \hline
\multicolumn{1}{|c|}{Dataset}                       & Loss Function                       & $\sigma_0$       & Train. Epochs               & Accuracy        & Avg. Spike Count    \\ \hline
\multicolumn{1}{|l|}{\multirow{4}{*}{CIFAR-10}}     & $CrossEntropy$ (baseline)           & 0                & 7                           & 64.97\%         & 294,130             \\ \cline{2-6} 
\multicolumn{1}{|l|}{}                              & $C.E. + \sigma_{constant}$          & 0.00055          & 19                          & 65.61\%         &  95,671             \\ \cline{2-6} 
\multicolumn{1}{|l|}{}                              & $C.E. + \sigma_{linear}$            & 0.00006          & 20                          & 65.30\%         &  95,544             \\ \cline{2-6} 
\multicolumn{1}{|l|}{}                              & $C.E. + \sigma_{quadratic}$         & 0.000008         & 15                          & 64.82\%         & 100,294             \\ \cline{2-6} 
\multicolumn{1}{|l|}{}                              & $C.E. + \sigma_{alternate}$         & 0.0015           & 20                          & 65.46\%         &  87,949             \\ \cline{2-6} 
\multicolumn{1}{|l|}{}                              & $C.E. + \sigma_{alternate\_linear}$ & 0.0004           & 16                          & 66.45\%         &  89,869             \\ \hline
\end{tabular}
\label{bestresults_strict}
\end{table}

\begin{table}[h]
\caption{Accuracy and spiking activity results using hyperparameters that give the best spiking sparsity with less than 1\% increase in validation error.}
\centering
\begin{tabular}{l|l|l|r|r|r|}
\cline{3-6}
\multicolumn{1}{c}{}                                &                                     & \multicolumn{2}{c|}{Best Val. Hyperparameters} & \multicolumn{2}{c|}{Testing Results} \\ \hline
\multicolumn{1}{|c|}{Dataset}                       & Loss Function                       & $\sigma_0$       & Train. Epochs               & Accuracy        & Avg. Spike Count   \\ \hline
\multicolumn{1}{|l|}{\multirow{4}{*}{CIFAR-10}}     & $CrossEntropy$ (baseline)           & 0                & 6                           & 64.19\%         & 276,586            \\ \cline{2-6} 
\multicolumn{1}{|l|}{}                              & $C.E. + \sigma_{constant}$          & 0.0007           & 20                          & 64.45\%         &  83,493            \\ \cline{2-6} 
\multicolumn{1}{|l|}{}                              & $C.E. + \sigma_{linear}$            & 0.0001           & 20                          & 64.03\%         &  81,902            \\ \cline{2-6} 
\multicolumn{1}{|l|}{}                              & $C.E. + \sigma_{quadratic}$         & 0.000008         & 19                          & 62.46\%         &  82,501            \\ \cline{2-6} 
\multicolumn{1}{|l|}{}                              & $C.E. + \sigma_{alternate}$         & 0.002            & 20                          & 63.90\%         &  78,411            \\ \cline{2-6}  
\multicolumn{1}{|l|}{}                              & $C.E. + \sigma_{alternate\_linear}$ & 0.00045          & 16                          & 65.29\%         &  83,864            \\ \hline
\end{tabular}
\label{bestresults}
\end{table}

\section{Discussion}

Discussion forthcoming. (Results are very preliminary and further experiments are underway.) Note that baseline accuracy values are low because these preliminary results are on LeNet and for the reduced training set (with the validation set removed). We expect that moving to better networks and re-including the removed validation set into training (after hyperparameter selection) will significantly improve results.

\section*{Conflict of Interest Statement}

The authors declare that the research was conducted in the absence of any commercial or financial relationships that could be construed as a potential conflict of interest.

\section*{Author Contributions}

JA wrote the paper. SS and JA performed the simulations. All authors assisted in developing the concepts.

\section*{Funding}
This work was supported in part by C-BRIC, a JUMP center sponsored by the Semiconductor Research Corporation and DARPA, and by the National Science Foundation, Intel Corporation, and the Vannevar Bush Fellowship.


\section*{Data Availability Statement}
The CIFAR dataset used in this study can be found at \href{https://www.cs.toronto.edu/~kriz/cifar.html}{[https://www.cs.toronto.edu/~kriz/cifar.html]}.

\bibliographystyle{frontiersinSCNS_ENG_HUMS}
\bibliography{SpikingSparsity}

\end{document}